\documentclass{article}
\usepackage{spconf,amsmath,graphicx,hyperref}
\usepackage{algpseudocode}
\usepackage{amsmath}
\usepackage{amssymb}
\usepackage{subfigure}
\usepackage{booktabs}
\usepackage{tabu}
\usepackage{bm}
\usepackage{multirow}
\usepackage{color}
\usepackage{amsmath,mathtools}
\usepackage[abs]{overpic}
\usepackage{makecell}   
\usepackage[table]{xcolor}

\title{Fusing in 3D: Free-Viewpoint Fusion Rendering \\ with a 3D Infrared-Visible Scene Representation}
\name{Chao Yang*, Deshui Miao*, Chao Tian, Guoqing Zhu, Yameng Gu and Zhenyu He**
\thanks{* Equal contribution
}
\thanks{** Corresponding Author
}
\thanks{Email: \{20b951014, 22b951002, tianchao, 20b951002, 23b951001\}@stu.hit.edu.cn, zhenyuhe@hit.edu.cn.}
}

\address{School of Computer Science and Technology, Harbin Institute of Technology, Shenzhen}
\begin{document}
\ninept
\maketitle
\begin{abstract}
Infrared-visible image fusion aims to integrate infrared and visible information into a single fused image. Existing 2D fusion methods focus on fusing images from fixed camera viewpoints, neglecting a comprehensive understanding of complex scenarios, which results in the loss of critical information about the scene. To address this limitation, we propose a novel Infrared-Visible Gaussian Fusion (IVGF) framework, which reconstructs scene geometry from multimodal 2D inputs and enables direct rendering of fused images.
Specifically,  we propose a cross-modal adjustment (CMA) module that modulates the opacity of Gaussians to solve the problem of cross-modal conflicts. Moreover, to preserve the distinctive features from both modalities, we introduce a fusion loss that guides the optimization of CMA, thus ensuring that the fused image retains the critical characteristics of each modality.
Comprehensive qualitative and quantitative experiments demonstrate the effectiveness of the proposed method.
\end{abstract}

\section{Introduction}
%
Infrared-visible image fusion merges thermal radiation information from infrared images with textural details from visible images to produce a comprehensive fused representation, leveraging their complementary characteristics under varying illumination conditions.
The fused results significantly benefit various computer vision tasks, including object detection \cite{yin2022vit,tian2024cross}, tracking \cite{cho2024flowtrack,wang2022mfgnet}, and semantic segmentation \cite{kirillov2023segment,lv2024context}. 
However, traditional 2D image fusion methods are limited to pixel-level integration from fixed camera perspectives, which constrains their applicability in scene understanding tasks.

\begin{figure}[t]
\centering
\includegraphics[width=0.95\linewidth]{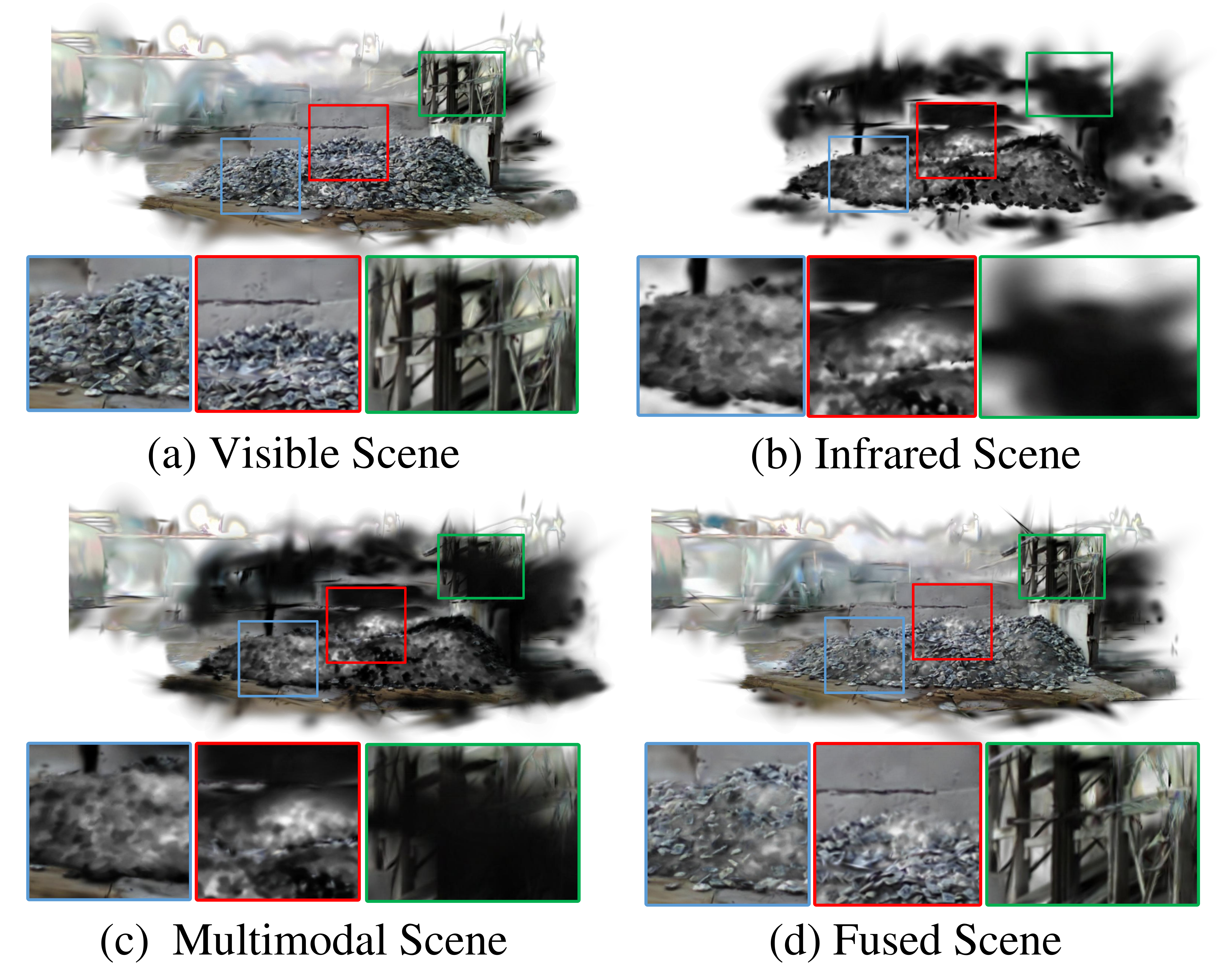}
\vspace{-0.4cm}
\caption{\textbf{3D Gaussian scene representation}. (a) Visible 3D scene. (b) Infrared 3D scene. (c) Multimodal 3D scene. (d) Our Infrared-Visible Gaussian Fusion (IVGF) scene. Direct rendering from multimodal scenes introduces cumulative errors, leading to the loss of detail and infrared information.}
\label{fig:first}
\end{figure}

Nowadays, multimodal scene representation introduces a novel paradigm that enables free-viewpoint rendering through multi-view 2D inputs, providing a more comprehensive representation of scene information. ThermalGaussian~\cite{lu2024thermalgaussian} introduced a 3DGS-based~\cite{3DGS} scene representation for infrared and visible modalities. 
MMOne~\cite{gu2025mmone} proposed a Gaussian-based multimodal decoupled representation that splits Gaussians into modality-specific and shared features to capture distinct modal characteristics separately. 
MS-Splatting~\cite{meyer2025multi} introduced a multi-spectral view synthesis framework that replaces traditional spherical harmonics with a unified neural color representation.
However, current multimodal scene representation methods are limited to separate-modality rendering and lack the capability for rendering fused images. As shown in Figure~\ref{fig:first}(c), modality conflicts in multimodal 3DGS often cause cross-modal error accumulation during direct rendering, resulting in the loss of critical information from both modalities.
An alternative approach employs a render-then-fusion pipeline that first renders each modality separately before fusing them, but this method incurs high computational overhead.
\begin{figure*}
\begin{center}
\includegraphics[width=1.0\linewidth]{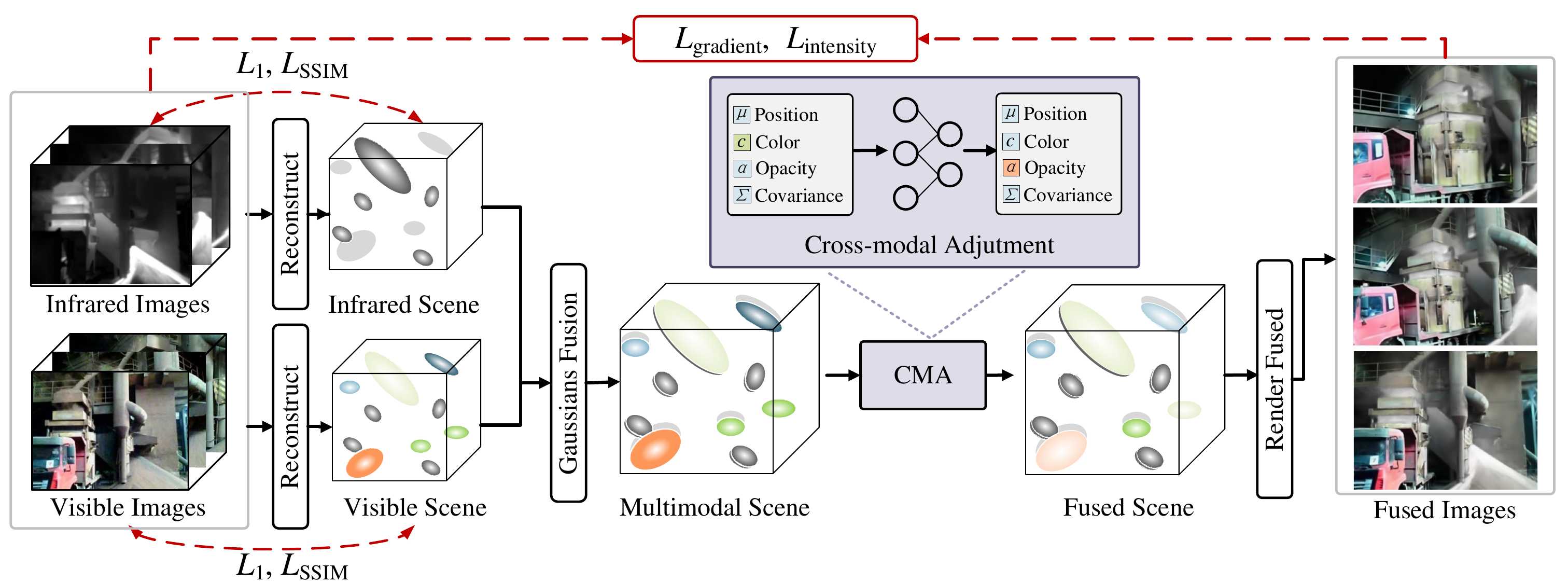}
\end{center}
\vspace{-0.6cm}
    \caption{\textbf{Overview of our Infrared-Visible Gaussian Fusion (IVGF) framework.} First, infrared and visible Gaussians are trained separately. Subsequently, they are concatenated and processed by a cross-modal adjustment (CMA) module, which refines the parameters of Gaussians to reconstruct a fused 3D scene.}

\label{fig:framework}
\vspace{-0.2cm}
\end{figure*}

To overcome these limitations, we propose Infrared-Visible Gaussian Fusion (IVGF), a novel 3D multimodal fusion framework that enables direct rendering of fused images from arbitrary viewpoints.
Specifically, to resolve modality conflicts during rendering, we introduce a cross-modal adjustment (CMA) module that modulates the opacity parameters of multimodal Gaussians based on their color attributes. 
Furthermore, to ensure the fused images retain salient characteristics from both modalities, we design a specialized fusion loss that preserves modality-specific features, thereby enhancing the overall rendering quality.
As demonstrated in Figure~\ref{fig:first}(d) (blue/red/green boxes), IVGF effectively preserves both infrared radiation characteristics and visible structural details, enabling comprehensive and informative scene characterization.

Our key contributions are summarized as follows:
\begin{itemize}
\item We propose Infrared-Visible Gaussian Fusion (IVGF), the first multimodal fusion representation framework that enables free-viewpoint rendering of fused images.

\item We introduce cross-modal adjustment (CMA) module that adjusts Gaussian opacity parameters to resolve modality conflicts during multimodal rendering.

\item We propose a fusion loss that preserves critical modality-specific features, which can obtain high rendering fidelity. 

\end{itemize}

\section{Methodology}
Traditional 2D image fusion methods are constrained by fixed camera viewpoints, thereby limiting their applicability in downstream tasks. To overcome this limitation, we propose a novel framework that reconstructs a fused infrared-visible 3D scene representation, enabling the rendering of fused images from arbitrary viewpoints.

\begin{figure*}
\begin{center}
\includegraphics[width=1.0\linewidth]{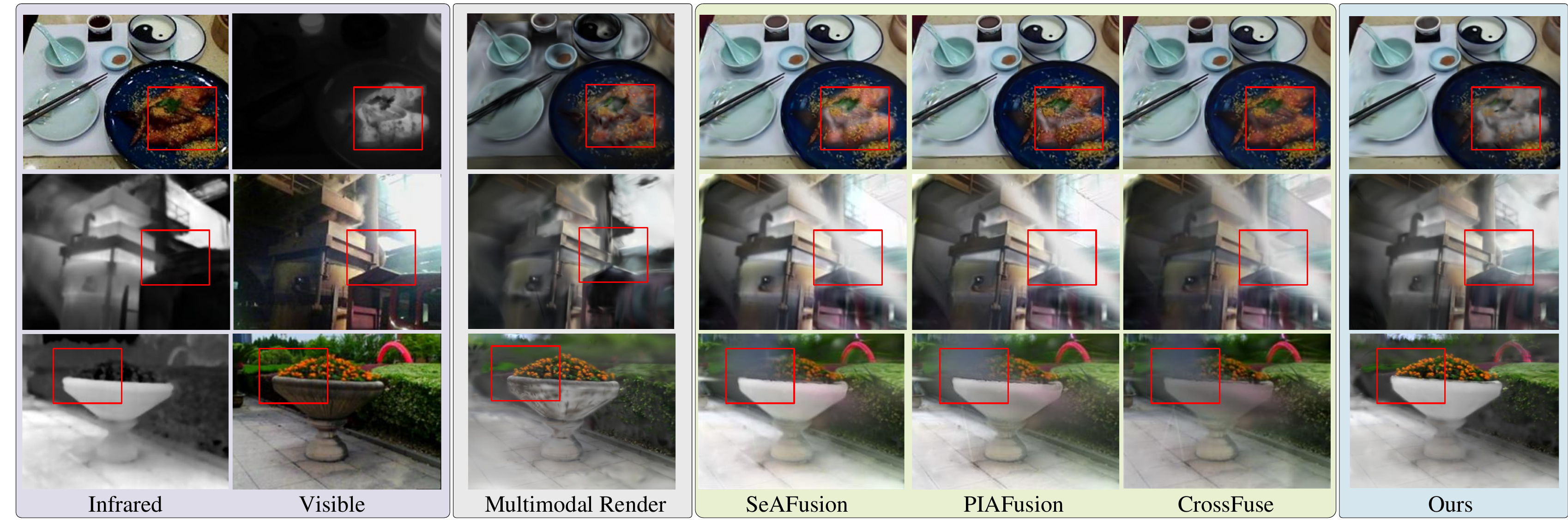}
\end{center}
\vspace{-0.6cm}
   \caption{\textbf{Qualitative results of different methods.} The first box contains infrared and visible images, the second box shows the results of multimodal rendering, the third box presents the fusion results from the render-then-fusion methods, and the final box displays our results.}

\label{fig:scene_compare}
\end{figure*}

\vspace{-0.2cm}
\subsection{Preliminaries: 3D Gaussian Splatting}
3D Gaussian Splatting (3DGS) framework \cite{3DGS}, which provides an efficient differentiable representation for radiance fields. The scene is represented as a collection \(\mathcal{G}=\{g_i\}_{i=1}^{N}\) of anisotropic 3D Gaussians, where each Gaussian primitive \(g_i\) represents a point in the scene and is parameterized by several essential attributes. The mean position \(\bm{\mu}_i \in \mathbb{R}^3\) defines the spatial location of each Gaussian in 3D space, while the covariance matrix \(\bm{\Sigma}_i = \mathbf{R}_i \mathbf{S}_i \mathbf{S}_i^\top \mathbf{R}_i^\top \in \mathbb{R}^{3 \times 3}\) controls its anisotropic scaling through the scaling matrix \(\mathbf{S}_i\) and its orientation via the rotation matrix \(\mathbf{R}_i\). Each Gaussian also possesses an opacity value \(\alpha_i \in [0,1]\) that determines its contribution to the final rendered image, along with view-dependent color properties \(\mathbf{c}_i\) encoded through spherical harmonics coefficients.

\vspace{-0.2cm}
\subsection{Infrared-Visible Gaussian Representation}
Due to the modality conflicts, direct rendering of fused images using multimodal 3DGS remains challenging. As demonstrated in Figure~\ref{fig:scene_compare} (c), rendered results often suffer from significant information loss, in which key structural details are obscured and critical infrared signatures are inadequately preserved.
To address this challenge, we propose a novel framework for 3DGS-based fused scene representation as shown in Figure~\ref{fig:framework}. The pipeline first constructs dual-modal scene representations, then optimizes the cross-modal adjustment (CMA) module to adjust opacity parameter to resolve modality conflicts during rendering.
The multimodal fusion rendering $F(\mathbf{x})$ at coordinate $\mathbf{x}\in\mathbb{R}^2$ is computed as follows:
\begin{equation}
F(\mathbf{x})=\sum_{k=1}^{N+M}c_k\,(\tau_k\alpha_k)\prod_{j=1}^{k-1}\!\bigl(1-\tau_j\alpha_j\bigr),
\end{equation}
where $N$ and $M$ denotes visible and infrared Gaussians contribute to the pixel, $c_k$ represents the color of the $k$-th Gaussian, $\alpha_k$ is its opacity, and $\tau_k$ denotes the cross-modal attention weight predicted by the cross-modal adjustment (CMA) module. 
CMA adjusts the opacities of infrared and visible Gaussians through a learned modality-aware MLP mechanism.
Specifically, the spherical harmonics coefficients $c_{\text{inf}}$ and $c_{\text{vis}}$ of the two Gaussians are concatenated and fed to MLP to predict $\tau$:

\begin{equation}
\tau = \text{MLP}(\text{Concat}(c_{\text{inf}}, c_{\text{vis}})),
\end{equation}
where $\tau \in [0,1]$ dynamically scales the opacity parameter of each Gaussian and $c \in \mathbb{R}^{(N+M) \times d_c}$, with $d_c$ denoting the dimension of the spherical harmonics coefficients per Gaussian. The MLP is calculated as follows:

\begin{equation}
\text{MLP}(\mathbf{C}) = \sigma\Big(
    \mathbf{W}_3^\top \cdot \phi_2\big(
        \phi_1(\mathbf{C})
    \big)
\Big),
\end{equation}
\begin{equation}
\phi_i(\mathbf{x}) = \text{LayerNorm}\big(\text{LeakyReLU}(\mathbf{x}\mathbf{W}_i + \mathbf{b}_i)\big) \quad (i=1,2),
\end{equation}
where $\mathbf{W}$ are weight matrices, $\sigma$ is the output activation function, and $\phi$ includes LeakyReLU activation and layer normalization.


\vspace{-0.2cm}
\subsection{Loss Function}
We adapt a two-stage training strategy for Infrared-Visible Gaussian Fusion (IVGF). In the first stage, 3D representation of the two modalities are trained using the same set up of ThermalGaussian~\cite{lu2024thermalgaussian}. In the second stage, two Gaussians are concatenated, and the cross-modal adjustment (CMA) module is trained to obtain the fused Gaussian representation.

\noindent\textbf{Training Stage 1.}
Following ThermalGaussian~\cite{lu2024thermalgaussian}, the Gaussian representations of the two modalities are trained separately. The loss function is formulated as:
\begin{equation}
\mathcal{L}_{\text{stage 1}}
= \gamma\,\mathcal{L}_{\text{Visible}} + (1-\gamma)\,\mathcal{L}_{\text{Infrared}},
\end{equation}
\begin{align}
\mathcal{L}_{\text{Visible}} &= \lVert V - V_{\text{render}}\rVert_1 + \bigl(1 - \text{SSIM}(V, V_{\text{render}})\bigr), \\[4pt]
\mathcal{L}_{\text{Infrared}} &= \lVert T - T_{\text{render}}\rVert_1 + \bigl(1 - \text{SSIM}(T, T_{\text{render}})\bigr).
\end{align}
Here V denotes visible image, T denotes thermal infrared image, and $\lVert\cdot\rVert_1$ denotes the $\ell_1$ distance, $\operatorname{SSIM}(\cdot,\cdot)$ denotes the structural similarity index, and the coefficient $\gamma=N_{\text{Infrared}}/(N_{\text{Infrared}}+N_{\text{Visible}})$ balances the contributions of the visible and infrared modalities, where $N_{\text{Infrared}}$ and $N_{\text{Visible}}$ denote the numbers of Gaussians in the corresponding modalities.

\noindent\textbf{Training Stage 2.}
To optimize a multimodal fused scene that preserves salient information from both modalities, we propose a novel fusion loss function to guide the cross-modal adjustment (CMA) module learning. Let $I_{\text{fuse}}$ denote the fused image rendered by the 3D fusion scene. The stage 2 cross-modal fusion loss function is formulated as:
\begin{equation}
\mathcal{L}_{\text{stage 2}} = \mathcal{L}_{\text{intensity}} + \mathcal{L}_{\text{gradient}},
\end{equation}
where $\mathcal{L}_{\text{intensity}}$ combines pixel-level similarity and structural similarity measures.

\begin{equation}
\label{eq:fusionloss}
\begin{multlined}
\mathcal{L}_{\text{intensity}} = \lambda_1\lVert I_{\text{fuse}} - \max(V, T) \rVert_1 \\
+ \lambda_2\bigl(1 - \text{SSIM}(I_{\text{fuse}},T)+(1 - \text{SSIM}(I_{\text{fuse}},V)\bigr),
\end{multlined}
\end{equation}
and $\mathcal{L}_{\text{gradient}}$ promotes edge retention between the two modalities, which is defined as follows:
\begin{equation}
\mathcal{L}_{\text{gradient}} = \lVert \nabla I_{\text{fuse}} - \nabla \max(V, T) \rVert_1,
\end{equation}
where $\nabla$ denotes the spatial gradient operator. 
The weighting parameters $\lambda_1$ and $\lambda_2$ balance between intensity preservation and structural maintenance. 
The proposed loss function helps preserve the most important information, thereby enhancing the representation capability of the fused multimodal scene.

\section{Experiments}
%
\subsection{Experimental Setup}
\noindent{\textbf{Datasets}}.
RGBT-Scenes dataset~\cite{lu2024thermalgaussian} contains over 1,000 aligned visible and thermal infrared image pairs. The dataset covers 10 diverse categories of scenes, including indoor and outdoor environments.
\noindent{\textbf{Evaluation Metrics}}.
%
Following the evaluation methodology for 2D fusion performance, we first calculate metrics between the fused image and the two source images, then average the metrics to assess overall performance. The metrics used for evaluation include: Peak Signal-to-Noise Ratio (PSNR), Structural Similarity Index (SSIM), and Learned Perceptual Image Patch Similarity (LPIPS).

\noindent{\textbf{Implement Details}}.
Our IVGF framework is built upon ThermalGaussian~\cite{lu2024thermalgaussian}, adopting all experimental parameters from the base model. The training process consisted of 30k iterations (15k for separate reconstruction phase followed by 15k for CMA optimization) on an RTX 3090 Ti GPU, generating fused images at a resolution of $640 \times 480$. For the loss function in Equation~\ref{eq:fusionloss}, we set the weights to $\lambda_1 = 1$ and $\lambda_2 = 2$.

\noindent{\textbf{Compared methods}}.
We adapt a render-then-fusion pipeline to generate fused images for evaluation and comparison. Specifically, ThermalGaussian~\cite{lu2024thermalgaussian} first renders paired infrared and visible images, and the two rendered images are then fed into state-of-the-art fusion methods to obtain the fused results, including CDDFuse~\cite{cddfuse}, CrossFuse~\cite{li2024crossfuse}, EMMA~\cite{emma}, LRRNet~\cite{li2023lrrnet}, LiMFusion~\cite{qian2024limfusion}, PIAFusion~\cite{tang2022piafusion}, SHIP~\cite{ship}, SeAFusion~\cite{TANG2022SeAFusion}, and SwinFusion~\cite{ma2022swinfusion}.

\begin{table*}[h]
\centering
\scriptsize
\caption{Quantitative performance comparison with 2D fusion methods (best in \textcolor{red}{Red}, second best in \textcolor{blue}{Blue}). The ↑ arrow indicates that higher values are better, while the ↓ arrow indicates that lower values are better.}
\renewcommand{\arraystretch}{1.2} 
\setlength{\tabcolsep}{2.8mm}
\begin{tabular}{l l*{10}{c}}
\hline
 & Method & \makecell{Daily\\Stuff} & Dimsum & \makecell{Glass\\Cup} & \makecell{Iron\\Ingot} & \makecell{Land\\Scape} & Parterre & \makecell{Plant\\Equipment} & \makecell{Road\\Block} & \makecell{Rotary\\Kiln} & Truck \\
\cline{1-12}
\multirow{10}{*}{LPIPS ↓}
 & CDDFuse (CVPR 2023) & 0.462 & 0.392 & 0.384 & 0.420 & 0.505 & 0.429 & 0.380 & 0.399 & 0.359 & 0.387 \\
 & CrossFuse (IF 2024) & 0.452 & 0.391 & 0.409 & 0.408 & \textcolor{blue}{0.495} & \textcolor{blue}{0.408} & \textcolor{red}{0.368} & \textcolor{blue}{0.373} & \textcolor{blue}{0.362} & 0.381 \\
 & EMMA (CVPR 2024) & 0.475 & 0.406 & 0.414 & 0.428 & 0.535 & 0.440 & 0.389 & 0.412 & 0.405 & 0.392 \\
 & LRRNet (TPAMI 2023) & \textcolor{blue}{0.448} & \textcolor{blue}{0.380} & \textcolor{blue}{0.375} & 0.422 & 0.512 & 0.417 & \textcolor{blue}{0.368} & 0.391 & \textcolor{red}{0.360} & \textcolor{blue}{0.370} \\
 & LiMFusion (OLEN 2024) & 0.494 & 0.428 & 0.456 & 0.427 & 0.528 & 0.441 & 0.396 & 0.424 & 0.401 & 0.407 \\
 & PIAFusion (IF 2022) & 0.453 & 0.393 & 0.381 & 0.416 & 0.515 & 0.426 & 0.387 & 0.378 & 0.391 & 0.388 \\
 & SHIP (CVPR 2024) & 0.525 & 0.445 & 0.398 & 0.463 & 0.604 & 0.522 & 0.483 & 0.460 & 0.517 & 0.456 \\
 & SeAFusion (IF 2022) & 0.452 & 0.389 & 0.380 & \textcolor{blue}{0.415} & 0.521 & 0.423 & 0.378 & 0.384 & 0.378 & 0.381 \\
 & SwinFusion (JAS 2022) & 0.459 & 0.385 & 0.376 & 0.418 & 0.517 & 0.428 & 0.377 & 0.396 & 0.367 & 0.382 \\
 \rowcolor{gray!20}
 & Ours & \textcolor{red}{0.425} & \textcolor{red}{0.378} & \textcolor{red}{0.371} & \textcolor{red}{0.413} & \textcolor{red}{0.481} & \textcolor{red}{0.395} & 0.370 & \textcolor{red}{0.366} & 0.369 & \textcolor{red}{0.372} \\
\hline

\multirow{10}{*}{PSNR ↑}  
 & CDDFuse (CVPR 2023) & 13.275 & 12.495 & 12.861 & 13.786 & 13.966 & 15.112 & 14.059 & 15.105 & 13.755 & 12.340 \\
 & CrossFuse (IF 2024) & 14.260 & 14.018 & 12.711 & \textcolor{red}{15.134} & \textcolor{blue}{15.056} & 16.531 & \textcolor{blue}{15.549} & \textcolor{red}{15.945} & \textcolor{red}{15.168} & 14.278 \\
 & EMMA (CVPR 2024) & 13.519 & 12.438 & 12.718 & 14.042 & 14.395 & 14.832 & 14.506 & 14.972 & 13.916 & 12.888 \\
 & LRRNet (TPAMI 2023) & \textcolor{red}{14.469} & \textcolor{blue}{13.590} & \textcolor{red}{14.197} & \textcolor{blue}{14.597} & 14.630 & \textcolor{blue}{15.814} & \textcolor{blue}{15.624} & 14.765 & \textcolor{blue}{14.839} & \textcolor{blue}{14.174} \\
 & LiMFusion (OLEN 2024) & 12.997 & 12.536 & 12.877 & 14.050 & 14.055 & 15.432 & 14.303 & 14.547 & 12.895 & 12.472 \\
 & PIAFusion (IF 2022) & 13.947 & 13.507 & \textcolor{blue}{13.918} & 14.771 & 13.756 & 15.659 & 14.407 & 15.084 & 12.027 & 12.893 \\
 & SHIP (CVPR 2024) & 9.073 & 9.608 & 12.569 & 10.412 & 6.116 & 9.397 & 8.878 & 9.072 & 5.859 & 9.485 \\
 & SeAFusion (IF 2022) & 13.127 & 12.812 & 13.473 & 13.878 & 13.706 & 14.677 & 13.826 & 14.049 & 12.852 & 12.443 \\
 & SwinFusion (JAS 2022) & 12.976 & 12.573 & 13.205 & 13.944 & 13.702 & 14.320 & 13.701 & 14.857 & 12.368 & 12.386 \\
 \rowcolor{gray!20}
 & Ours & \textcolor{blue}{13.929} & \textcolor{red}{14.738} & 12.968 & 14.553 & \textcolor{red}{15.579} & \textcolor{red}{16.646} & 15.047 & \textcolor{red}{15.620} & 12.592 & \textcolor{red}{13.152} \\
\hline

\multirow{10}{*}{SSIM ↑} 
 & CDDFuse (CVPR 2023) & 0.614 & 0.541 & 0.577 & 0.490 & \textcolor{blue}{0.572} & 0.696 & 0.657 & 0.808 & 0.575 & 0.573 \\
 & CrossFuse (IF 2024) & 0.613 & 0.544 & 0.570 & 0.490 & 0.558 & \textcolor{blue}{0.707} & 0.661 & \textcolor{blue}{0.831} & 0.552 & 0.569 \\
 & EMMA (CVPR 2024) & 0.602 & 0.545 & 0.575 & 0.489 & 0.548 & 0.685 & 0.652 & 0.805 & 0.559 & \textcolor{blue}{0.590} \\
 & LRRNet (TPAMI 2023) & \textcolor{blue}{0.630} & 0.540 & 0.592 & 0.484 & 0.552 & 0.682 & 0.648 & 0.802 & 0.553 & 0.546 \\
 & LiMFusion (OLEN 2024) & 0.601 & 0.535 & 0.570 & 0.484 & 0.540 & 0.681 & 0.648 & 0.792 & 0.561 & 0.559 \\
 & PIAFusion (IF 2022) & 0.615 & 0.548 & 0.593 & 0.489 & 0.531 & 0.695 & 0.647 & 0.812 & 0.557 & 0.573 \\
 & SHIP (CVPR 2024) & 0.526 & 0.484 & 0.568 & 0.442 & 0.428 & 0.591 & 0.531 & 0.729 & 0.413 & 0.491 \\
 & SeAFusion (IF 2022) & 0.619 & \textcolor{blue}{0.555} & \textcolor{blue}{0.596} & \textcolor{blue}{0.495} & 0.559 & 0.703 & \textcolor{blue}{0.662} & 0.812 & 0.582 & 0.589 \\
 & SwinFusion (JAS 2022) & 0.616 & 0.553 & 0.591 & 0.494 & 0.559 & 0.692 & 0.661 & 0.807 & \textcolor{blue}{0.586} & 0.586 \\
 \rowcolor{gray!20}
 & Ours & \textcolor{red}{0.665} & \textcolor{red}{0.567} & \textcolor{red}{0.613} & \textcolor{red}{0.495} & \textcolor{red}{0.610} & \textcolor{red}{0.734} & \textcolor{red}{0.680} & \textcolor{red}{0.839} & \textcolor{red}{0.618} & \textcolor{red}{0.603} \\
\hline
\end{tabular}
\label{tab:lpips_results}
\vspace{-0.3cm}
\end{table*}

\subsection{Qualitative Analysis of Fusion Methods}
Figure~\ref{fig:scene_compare} compares the results of our method with direct multimodal 3DGS rendering results and other 2D fusion methods. 
Rendering of multimodal 3DGS result leads to an accumulation of errors, resulting in blurry images and the loss of critical information. 
For 2D fusion methods (render-then-fusion strategy), the regions within the red bounding boxes tend to lose critical thermal information and exhibit blurring as well as abrupt illumination changes. 
In contrast, our method effectively preserves the thermal information from the infrared modality and the textural details from the visible modality, avoiding these artifacts. 
These comparisons demonstrate that our proposed IVGF effectively preserves and integrates information from both modalities, achieving a more comprehensive multimodal fusion representation.

\subsection{Quantitative Analysis of Fusion Methods}
The quantitative comparisons in Table~\ref{tab:lpips_results} show that IVGF achieves the highest scores in SSIM and LPIPS across almost all scenes, outperforming nine existing fusion methods (red denotes the best performance, blue the second-best). The results demonstrate that our proposed IVGF can effectively reconstruct fused multimodal scenes and produce higher-quality fusion results than 2D fusion methods (render-then-fusion strategy).

\begin{figure}
\begin{center}
\includegraphics[width=0.8\linewidth]{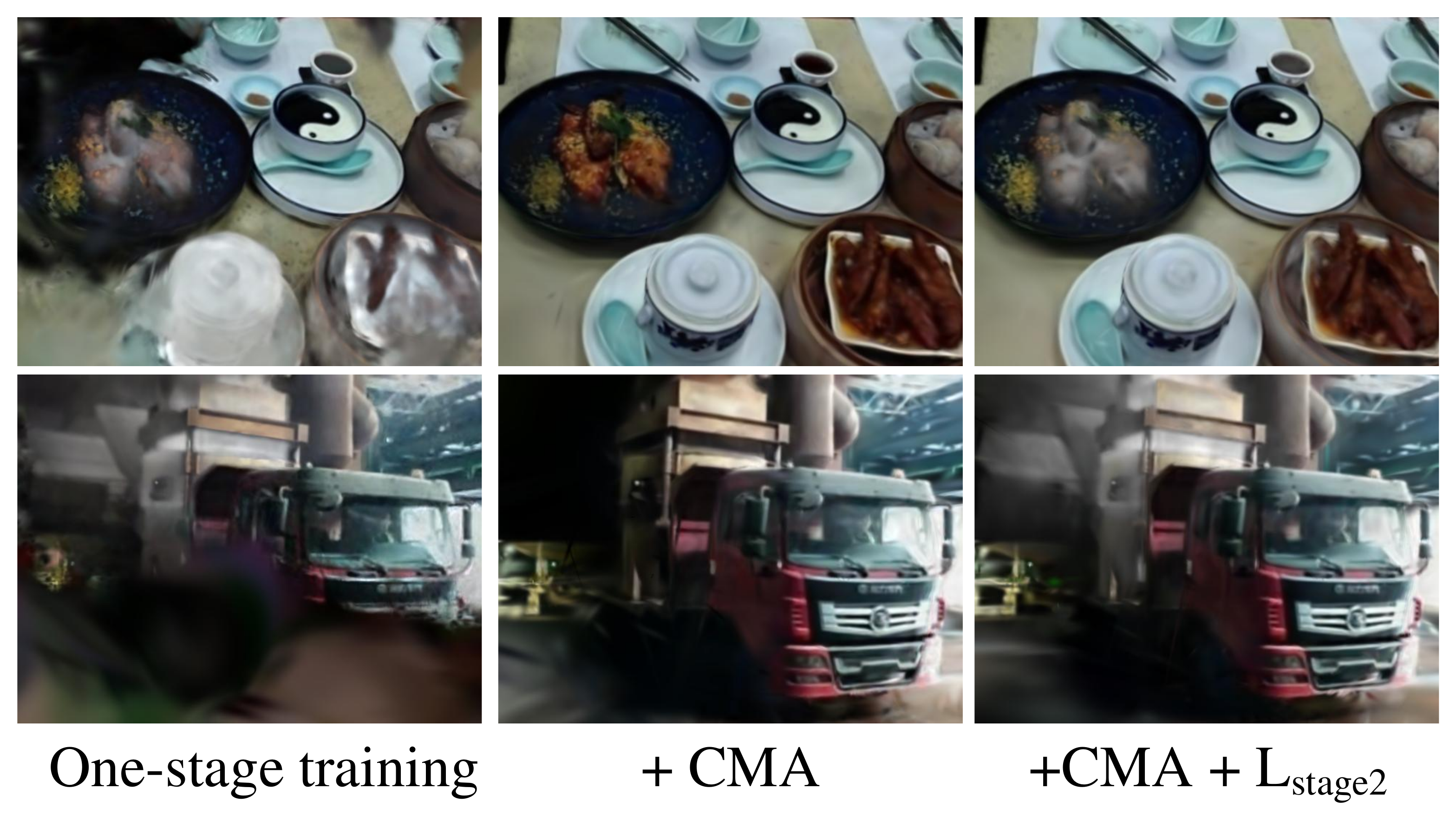}
\end{center}
\vspace{-0.6cm}
    \caption{\textbf{Ablation study on the effects of different modules.} The cross-modal adjustment (CMA) module resolves cross-modal error accumulations, while the proposed fusion loss ensures the retention of essential features from both modalities in the rendered results.}

\label{fig:ablation}
\vspace{-0.5cm}
\end{figure}

\begin{table}[h]
\vspace{-0.3cm}
\renewcommand{\arraystretch}{1.2}
\scriptsize
\centering
\caption{Ablation study of our method. (Best in \textcolor{red}{Red}, Second Best in \textcolor{blue}{Blue}). The $\uparrow$ arrow indicates that higher values are better, while the $\downarrow$ arrow indicates that lower values are better.}
\label{tab:comparison}
\resizebox{1.0\linewidth}{!}{
\begin{tabular}{lc|ccccc}
\toprule
\textbf{} & \textbf{Methods} & \textbf{LPIPS $\downarrow$} & \textbf{PSNR $\uparrow$} & \textbf{SSIM $\uparrow$} & \textbf{Mem. $\downarrow$} & \textbf{FPS $\uparrow$} \\
\midrule
\multirow{2}{*}{1} & Multimodal Render & 0.408 & 12.426 & 0.515 & \color{red}{20 MB} & \color{red}{631} \\
 & 2D Fusion & 0.389 & 12.812 & 0.555 & 69 MB & 239 \\
\midrule
\multirow{3}{*}{2} & One-stage training & 0.503 & 10.735 & 0.455 & 376 MB & 138 \\
 & + CMA & \color{blue}{0.371} & \color{blue}{14.660} & \color{blue}{0.567} & 23 MB & 387 \\
\rowcolor{gray!20}
 & + CMA + $L_{\text{stage 2}}$ & \color{red}{0.378} & \color{red}{14.738} & \color{red}{0.567} & \color{blue}{21 MB} & \color{blue}{412} \\
\bottomrule
\end{tabular}}
\vspace{-0.6cm}
\end{table}

\subsection{Ablation Study}
\noindent\textbf{Comparison with different 3D fusion strategies.} As shown in Figure~\ref{fig:scene_compare} and Table~\ref{tab:comparison}.1, our method achieves a more favorable balance among visual quality, storage efficiency, and rendering speed.
Specifically, IVGF outperforms multimodal rendering in all performance metrics while matching its memory and speed. Compared to the 2D render-then-fusion strategy, IVGF achieves superior performance with higher PSNR and SSIM, lower LPIPS, a faster speed of 182 FPS, and a reduced memory usage of 48 MB. These results demonstrate that IVGF establishes an effective and efficient new paradigm for 3D scene representation.

\noindent{\textbf{Effect of different modules}}.
As shown in Figure~\ref{fig:ablation} and Table~\ref{tab:comparison}.2, ablation study shows that the cross-modal adjustment (CMA) module enhances structural details, leading to improvements in visual quality, evaluation metrics, processing speed, and memory efficiency. This improvement stems from the CMA module's ability to effectively resolve conflicts between multimodal Gaussians, thereby achieving higher performance than one-stage training strategy.
Moreover, with the proposed fusion loss, visual results exhibit enhanced retention of salient information from both modalities, while quantitative evaluations confirm consistent improvements across all metrics. This demonstrates that our proposed loss function effectively preserves critical features from each modality within the fused multimodal scene representation.

%

\section{Conclusion}
In this work, we present Infrared-Visible Gaussian Fusion (IVGF), the first framework to achieve free-viewpoint rendering of fused images through a multimodal fusion 3D representation. To this end, we design a cross-modal adjustment (CMA) module to resolve modality conflicts by dynamically modulating Gaussian opacity and introduce a novel fusion loss to retain salient information from two modalities. Extensive experiments validate the effectiveness of our approach.

\bibliographystyle{IEEEbib}
\bibliography{strings,refs}

\end{document}